\documentclass{article}

\usepackage[preprint]{sty/neurips_2020}

\setcitestyle{numbers}
\usepackage[utf8]{inputenc} 
\usepackage[T1]{fontenc}    
\usepackage{hyperref}       
\usepackage{url}            
\usepackage{amsfonts}       
\usepackage{nicefrac}       
\usepackage{microtype}      

\usepackage{times}
\usepackage{epsfig}
\usepackage{graphicx}
\usepackage{amsmath}
\usepackage{amssymb}
\usepackage{paralist}
\usepackage{multirow}
\usepackage{colortbl}

\usepackage{booktabs}
\usepackage{sty/mysymbols}
\usepackage{sty/widetext}
\usepackage{enumitem}
\usepackage[ruled,vlined,linesnumbered]{algorithm2e}
\usepackage[table]{xcolor}
\usepackage{float}
\usepackage{hyperref}

\usepackage{sty/space_saver}

\newcommand{\stageone}{Zero-shot Transfer}
\newcommand{\ours}{Ours}
\newcommand{\nonvar}{Typical Transfer}
\renewcommand{\paragraph}[1]{\vspace{0pt}\noindent\textbf{#1}}

\date{}

\title{Dialog without Dialog Data: Learning Visual Dialog Agents from VQA Data}

\author{Michael Cogswell$^5$\thanks{Equal contribution. Work carried out mainly at Georgia Tech.} \enskip Jiasen Lu$^3$\footnotemark[1] \enskip Rishabh Jain $^{1}$ \enskip Stefan Lee$^{2}$ \enskip Devi Parikh$^{4,1}$ \enskip Dhruv Batra$^{4,1}$ \\
$^1$Georgia Institute of Technology \quad $^2$Oregon State University \\ \quad$^3$ Allen Institute for AI \quad$^4$ Facebook AI Research \quad$^5$ SRI International\\
\tt\small michael.cogswell@sri.com \tt\small jiasenl@allenai.org \tt\small leestef@oregonstate.edu \\
\tt\small \{rishabhjain, parikh, dbatra\}@gatech.edu}

\newcommand{\suppSectionArchitectureDetails}{2}
\newcommand{\suppSectionHumanStudies}{3}
\newcommand{\suppSectionAblations}{5}
\newcommand{\suppSectionRelWork}{6}

\begin{document}
\maketitle
\begin{abstract}

Can we develop visually grounded dialog agents that can efficiently adapt to new tasks without
forgetting how to talk to people?
Such agents could leverage a larger variety of existing data to generalize to new task, 
minimizing expensive data collection and annotation.
In this work, we study a setting we call ``\textit{Dialog without Dialog}'', 
which requires agents
to develop visually grounded dialog models that can adapt to new tasks without
language level supervision.
By factorizing intention and language, our model minimizes
linguistic drift after fine-tuning for new tasks.
We present qualitative results, automated metrics, and human studies that
all show our model can adapt to new tasks and maintain language quality.
Baselines either fail to perform well at new tasks or experience language drift,
becoming unintelligible to humans. Code has been made available at: \href{https://github.com/mcogswell/dialog_without_dialog}{https://github.com/mcogswell/dialog\_without\_dialog}

\end{abstract}

\section{Introduction}
One goal of AI is to enable humans and computers to
communicate naturally with each other in grounded language
to achieve a collaborative objective. Recently the community has studied 
goal oriented dialog, where agents communicate
for tasks like booking a flight or searching for images \cite{miller2017parlai}.

A popular approach to these tasks has been to observe humans engaging 
in dialogs like the ones we would like to automate and then train agents 
to mimic these human dialogs~\cite{visdial_rl,deal_or_no_deal}.
Mimicking human dialogs allows agents to generate intelligible language (\ie, 
meaningful English, not gibberish). However, these models are typically fragile and generalize poorly to new tasks. 
As such, each new task requires collecting new human dialogs, which is a laborious 
and costly process often requiring many iterations
before high quality dialogs are elicited \cite{visdial, guesswhat}. 

A promising alternative is to use goal completion as a supervisory signal
to adapt agents to new tasks.
Specifically, this is realized by pre-training dialog agents via human dialog supervision on one
task and then fine-tuning them on a new task by rewarding the agents
for solving the task regardless of the dialog's content. This approach can indeed 
improve task performance, but language quality suffers 
even for similar tasks. It tends to drifts from human language, becoming ungrammatical 
and loosing human intelligible semantics -- sometimes even turning into unintelligible code.
Such code may allow communication with other bots, but is largely incomprehensible to humans. 
This trade off between task performance and language drift has been observed in prior dialog work~\cite{visdial_rl,deal_or_no_deal}.
 
\begin{figure}
    \centering
    \includegraphics[width=\textwidth]{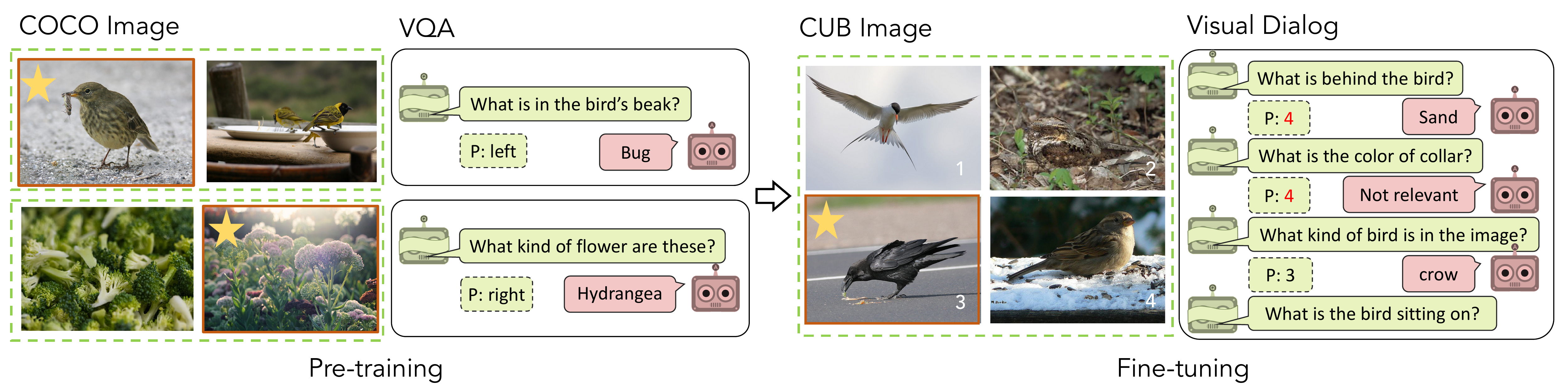}
    \vspace{-15pt}
    \caption{An example of \textit{Dialog without Dialog} (DwD) task: (Left) We pre-train questioner agent (\qbot{} in green) that can discriminate
    between pairs of images by mimicking questions from VQAv2 \cite{vqav2}. 
    (Right) \qbot{} needs to generate a sequence of discriminative questions (a dialog) to identify the secret image that \abot{} picked. Note that the language supervision is not available, thus we can only fine-tune \qbot{} with task performance. In DwD, we can test \qbot{} generalization ability by varying dialog length, pool size and image domain.
    }
    \vspace{-10pt}
    \label{fig:guesswhich}
\end{figure}
 
The goal of this paper is to 
develop visually grounded dialog models that can adapt to new tasks while exhibiting less linguistic drift, thereby reducing
the need to 
collect new data for the new tasks.
To test this we consider an image guessing game demonstrated in \figref{fig:guesswhich} (right).
In each episode, one agent (\abot{} in red) secretly selects a target image $y$ (starred) from a pool of images. The other agent (\qbot{} in green) must identify this image by asking questions
for \abot{} to answer in a dialog. To succeed, \qbot{} needs to understand the
image pool, generate discriminative questions, and interpret the answers \abot{}
provides to identify %
the secret image. 
The image guessing game provides the agent with a goal, and we can test \qbot{} generalization ability by varying dialog length, pool size and image domain.


\paragraph{Contribution 1.}
We propose the \textit{Dialog without Dialog} (\dwd{}) task, which requires a \qbot{} to perform our image
guessing game without dialog level language supervision.
%
%
As shown in in \figref{fig:guesswhich} (left),  \qbot{} in this setting first learns to ask questions to identify the secret image by mimicking single-round human-annotated visual questions. For the dialog task (right), no human dialogs are available so \qbot{} can only be supervised by its image guessing performance. To measure task performance and language drift in increasingly out-of-distribution settings we consider varied pool sizes and take pool images from diverse image sources (\eg close-up bird images).



\paragraph{Contribution 2.}
We propose a \qbot{} architecture for the \dwd{} task
that decomposes question intent from the words used to express that intent.
%
We model the question intent by introducing a discrete latent representation 
that is the only input to the language decoder. 
We further pair this with a \textit{pre-train then fine-tune} learning approach that teaches \qbot{} how to ask visual questions from VQA during pre-training and `what to ask' during fine-tuning for visual dialogs.

\paragraph{Contribution 3.}
We measure \qbot{}'s ability to adapt to new tasks and maintain language
quality.
Task performance is measured with both automatic and human answerers
while language quality is measured using three automated metrics
and two human judgement based metrics.
Our results show the proposed \qbot{} both adapts to new tasks better than
a baseline chosen for language quality and maintains language quality better
than a baseline optimized for just task performance.


\section{Dialog Based Image Guessing Game}

\subsection{Game Definition}
\label{sec:guessgame}

Our image guessing game proceeds one round at a time, starting at round $r = 1$ and running
for a fixed number of rounds of dialog $R$.
At round $r$, \qbot{} observes the pool of images $\mathcal{I} =\{I_1, \dots, I_P\}$ of size $P$, 
the history of question answer pairs $q_1, a_1, \dots q_{r-1}, a_{r-1}$, and placeholder representations $q_0, a_0$
that provide input for the first round. It generates a question
\begin{equation}
q_{r} =
    \mathtt{QBot.Ask}(\mathcal{I}, q_0, a_0, \dots q_{r-1}, a_{r-1}).
\end{equation}
Given this question, but not the entire dialog history, \abot{} answers based on the randomly selected target image $I_{\hat{y}}$ (not known to \qbot{}):
\begin{equation}
a_{r} =
    \mathtt{ABot.Answer}(I_{\hat{y}}, q_{r}).
\end{equation}
Once \qbot{} receives the answer from \abot{}, it makes a prediction $y_{r}$ guessing the target image:
\begin{equation}
y_{r} =
    \mathtt{QBot.Predict}(\mathcal{I}, q_0, a_0, \dots , q_{r}, a_{r})
\end{equation}
%

\paragraph{Comparison to GuessWhich.}
Our Image Guessing game is inspired by GuessWhich game of \citet{visdial_rl}, and there are two subtle but important differences.
In GuessWhich, \qbot{} initially observes a caption describing \abot{}'s selected image and
must predict the selected image's features to retrieve it from a large, fixed pool of images it does not observe. 
First, the inclusion of the caption leaves little room for the dialog to add information~\cite{mironenco2017examining}, so we omit it.
Second, in our game a small pool of images is sampled for each dialog and \qbot{} directly predicts the target image given those choices.

\subsection{Modelling \abot{}}
\label{sec:abot}

In this work, we focus primarily on \qbot{} agent rather than \abot{}. 
We set \abot{} to be a standard visual question answering agent, specifically the Bottom-up Top-down \cite{bottomup_tricks} model; however, we do make one modification. \qbot{} may generate questions that are not well 
grounded in \abot{}'s selected image (though they may be grounded in other pool images) -- e.g. asking about a surfer when none exists.
To enable \abot{} to respond appropriately, we augment \abot{}'s answer space with a \texttt{Not Relevant} token. 
To generate training data for this token we augment every image with an additional randomly sampled question and
set \texttt{Not Relevant} as its target answer.
\abot{} is trained independently from \qbot{} on the VQAv2 dataset and then frozen.

\subsection{Modelling \qbot{}}
\label{sec:qbot}

We conceptualize \qbot{} as having three major modules.
The \emph{planner} encodes the state of the game to decide what to ask about.
The \emph{speaker} takes this intent and formulates the language to express it.
The \emph{predictor} makes target image predictions taking the dialog history into account.
We make fairly standard design choices here, then adapt this model for the \dwd{} task
in \secref{sec:dwd}.

\paragraph{Pool \& Image Encoding.}
We represent the $p$-th image $I_p$ of the pool as a set of $B$ bounding boxes such that 
$I^p_b$ is the embedding of the $b$-th box using the same Faster R-CNN~\cite{fasterrcnn} embeddings as
in \cite{bottomup}.
Note that we do not assume prior knowledge about the size or composition of the pool.

\begin{figure}[t]
    \centering
    \includegraphics[width=0.7\textwidth]{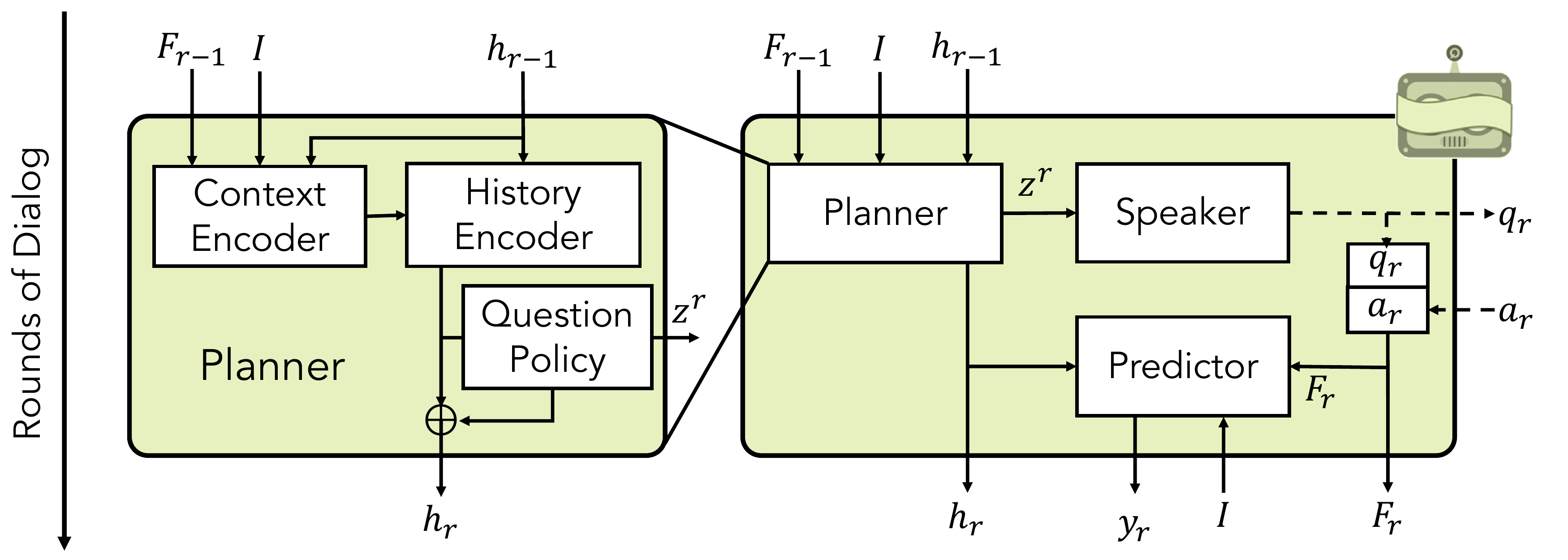}
    \caption{\footnotesize{A single round of our \qbot{} which decomposes into the modules described in \secref{sec:qbot}. This factorization
    allows us to limit language drift while fine-tuning for task performance.}}
    \label{fig:qbot}
    \vspace{-10pt}
\end{figure} 

\paragraph{Planner.}
The planner's role is to encode the dialog context (image pool and dialog history) into representation $z^r$,
deciding what to ask about in each round. It also produces an encoding $h_r$ of the dialog history.
To limit clutter, we denote the question-answer pair at round $r$ as a `fact' $F_r = [q_r, a_r]$. 

\paragraph{Planner -- Context Encoder.}
Given the prior dialog state $h_{r-1}$, $F_{r-1}$, and image
pool $\mathcal{I}$, the context encoder performs 
hierarchical attention over images in $\mathcal{I}$ and object boxes in each image to identify image regions
that are most relevant for generating the next question. 
As we detail in Section \suppSectionArchitectureDetails{} of the supplement, $F_{r-1}$ and $h_{r-1}$ query the image to attend to relevant regions across the pool.
First a $P$ dimensional distribution $\alpha$ over images in the pool is produced
and then a $B$ dimensional distribution $\beta^p$ over boxes is produced for each image $p \in \{1, \ldots, P\}$.
The image pool encoding $\hat{v}_r$ at round $r$ is
\begin{align}
    \hat{v}_r = \sum_{p=1}^P \alpha_p \sum_{b=1}^B \beta^p_b I^p_b.
\end{align}
This combines the levels of attention and is agnostic to pool size.

\paragraph{Planner -- History Encoder.} 
To track the state of the game, the planner applies an LSTM-based history encoder that takes $\hat{v}_r$ and $F_{r-1}$ as input and produces an intermediate hidden state $h_{r}$.
Here $h_{r}$ includes a compact representation of question intent and dialog history, helping provide a differentiable connection between the intent and final predictions through the dialog state. 

\paragraph{Planner -- Question Policy.}
The question policy transforms $h_{r}$ to this module's output $z^r$, which the speaker decodes into a question.
By default $z^r$ is equal to the hidden state $h_r$,
but in \secref{sec:new_qbot} we show how 
a discrete representations 
can be used to reduce language drift.

\paragraph{Speaker.}
Given an intent vector $z^r$, the speaker generates a natural language question. Our speaker is a standard LSTM-based decoder with an initial hidden state equal to $z^r$. 

\paragraph{Predictor.}
The predictor uses the dialog context generated so far to guess the target image. It takes a concatenation $F = [F_1, \ldots, F_{r}]$ of fact embeddings and the dialog
state $h_{r}$ and computes an attention pooled fact $\hat{F}$ using $h_{r}$ as attention context.
Along with $h_{r}$, this is used to attend to salient image features then compute a distribution over images in the pool
using a softmax (see Algorithm 2 in the supplement for full details), allowing for
the use of cross-entropy as the task loss.
Note that the whole model is agnostic to pool size.

\section{Dialog without Dialog}
\label{sec:dwd}

Aside from some abstracted details, the game setting and model presented in the previous section could be trained without any further information -- a pool of images could be generated, 
\abot{} could be assigned an image, the game could be rolled out for arbitrarily many rounds, and \qbot{} could be trained to predict the correct image given \abot{}'s answers. While 
this is an interesting research direction in its own right~\cite{ren20,chaabouni20,liang20}, there is an obvious shortcoming --
it would be highly improbable for \qbot{} to discover a fully functional language that humans can already understand.
Nobody discovers French. They have to learn it.

At the other extreme -- representing standard practice in dialog problems -- humans could be recruited to perform this image guessing game and provide dense supervision for what questions \qbot{} should ask to perform well at this specific task. However, this requires collecting language data for every new task. It is also intellectually dissatisfying for agents' knowledge of natural language to be so inseparably intertwined with individual tasks. After all, one of the greatest powers of language is the ability to use it to communicate about many different problems.

In this section, we consider a middle ground that has two stages. Stage 1 trains our agent on one task where training data already exists (VQA; \ie, single round dialog) and then 
stage 2 adapts it to carry out goal driven dialog (image guessing game) without further supervision. 

\subsection{Stage 1: Language Pre-training}
\label{sec:stage1}

We leverage the VQAv2~\cite{vqav2} dataset as our language source to 
learn how to ask questions that humans can understand. By construction, for each question in VQAv2 there exists at least one pair of images which are visually similar but have different ground truth answers to the same question.
Fortuitously, this resembles our dialog game -- the image pair is the pool, the question is guaranteed to be discriminative, and we can provide an answer depending on \abot{}'s selected image.
We view this as a special case of our game that is fully supervised but contains only a single round of dialog.
During stage 1 \qbot{} is trained to mimic the human question (via cross-entropy teacher forcing) and to predict the correct image given the ground truth answer.

For example, in the top left of \figref{fig:guesswhich} outlined in dashed green we show a pair of two bird images with the question ``What is in the bird's beak?'' from VQAv2.
Our agents engage in a single round dialog where \qbot{} asks that question and \abot{} provides the answer (also supervised by VQAv2).

\subsection{Stage 2: Transferring to Dialog}
\label{sec:new_qbot}

A first approach for adapting agents would be to
take the pre-trained weights from stage 1 and simply fine-tune for our full image guessing task.
However, this agent would face a number of challenges. It has never had to model multiple steps of a dialog.
Further, while trying to predict the target image there is little to encourage \qbot{} to continue producing intelligible language.
Indeed, we find our baselines do exhibit language drift.
We consider four modifications to address these problems.


\paragraph{Discrete Latent Intention Representation $z^r$.}
\label{sec:pre-training}
Rather than a continuous vector passing from the question policy to the speaker, we pass discrete vectors.
Specifically, we consider a representation composed of $N$ different $K$-way Concrete variables~\cite{concrete_distribution}.
Let $z^r_n \in \{1, \ldots, K\}$ and let the logits $l_{n,1}, \ldots, l_{n,K}$ paramterize the Concrete distribution $p(z^r_n)$.
We learn a linear transformation $W^z_n$ from the intermediate dialog state $h_{r}$ to produce these logits for each variable $n$:%
\begin{align}
    l_{n,k} &= \mathrm{LogSoftmax}\left(W^z_n h_{r} \right)_k \; \forall k \in \{1, \ldots, K\} \; \forall n \in \{1, \ldots, N\}
\end{align}
To provide input to the speaker, $z^r$ is embedded using a learned dictionary of embeddings.
In our case each variable in $z^r$ has a dictionary of $K$ learned embeddings.
The value of $z^r_n$ ($\in \{1, \ldots, K\}$) picks one of the embeddings
for each variable and the final representation simply sums over all variables:
\begin{align}
    e_z = \sum_{n=1}^{N} E_n(z^r_n).
\end{align}
\paragraph{VAE Pre-training.}
When using this representation for the intent, we train stage 1 by replacing the likelihood with an ELBO (Evidence Lower BOund) loss
as seen in Variational Auto-Encoders (VAEs)~\cite{vae} to
help disentangle intent from expression by
restricting information flow through $z^r$.
We use the existing speaker module to decode $z^r$ into questions and train a new encoder
module to encode ground truth VQAv2 question $\hat{q}_1$ into conditional distribution $q(z^1 | \hat{q}_1, \mathcal{I})$ over $z^r$ at round 1.

For the encoder we use a version of the previously described context encoder from \secref{sec:qbot} that
uses the question $\hat{q}_1$ as attention query instead of $F_{r-1}$ and $h_{r-1}$ (which are not available in this context).
The resulting ELBO loss is
\begin{align}
    \mathcal{L} = 
    & E_{z^1 \sim q(z^1 | \hat{q}_1, \mathcal{I})}\left[ \log p(\mathtt{speaker}(z^1)) \right] +
        \frac{1}{N} \sum_{n=1}^{N} D_{KL}\left[ q(z^1_n | \hat{q}_1, \mathcal{I}) || \mathcal{U}(K) \right]
    \label{eq:elbo}
\end{align}
This is like the Full ELBO described, but not implemented, in \cite{rethink_zhao}.
The first term encourages the encoder to represent and the speaker to mimic the VQA question.
The second term uses the KL Divergence $D_{KL}$ to push the distribution of $z$ close to
the $K$-way uniform prior $\mathcal{U}(K)$, encouraging $z$ to ignore irrelevant information.
Together, the first two terms form an ELBO on the question likelihood
given the image pool~\cite{gumbel_softmax,kaiser18}.

\paragraph{Fixed Speaker.}
Since the speaker contains only lower level
information about how to generate language, we freeze it during task transfer. We want only the
high level ideas represented by $z$ and the predictor which receives direct
feedback to adapt to the new task. If we updated the speaker then its language
could drift given only the sparse feedback available in each new setting.

\paragraph{Adaptation Curriculum.}
As the pre-trained (stage 1) model has never had to keep track of dialog contexts beyond the first round, we fine-tune in two stages, 2.A and 2.B. In {\bf stage 2.A} we fix the Context Encoder and Question Policy parts of the planner
so the model can learn to track dialog effectively without trying
to generate better dialog at the same time. This stage takes 20 epochs
to train. Once \qbot{} learns how to track dialog we update the
entire planner in {\bf stage 2.B} for 5 epochs.\footnotemark

\footnotetext{We find 5 epochs stops training early enough to
avoid overfitting on our val set.}
\section{Experiments} \label{sec:experiments}

We want to show that our proposed agent can adapt to new tasks while exhibiting less linguistic drift.
In \secref{sec:experimental_settings} and \secref{sec:baselines} we start by describing
the new tasks we construct and the baselines we compare to, then
the following sections demonstrate how our model adapts while preventing drift using
qualitative examples (\secref{sec:qualitative}),
automated metrics (\secref{sec:auto_eval}),
and human judgements (\secref{sec:human-study}).
We also summarize the model ablations (\secref{sec:ablations}) detailed in the supplement.

\paragraph{Task Settings.}
\label{sec:experimental_settings}
We construct new tasks by varying four parameters of our image guessing game:
\begin{compactitem}[--]
\item \textbf{Number of Dialog Rounds.}
The number of dialog rounds $R$ is fixed at 1, 5, or 9.

\item \textbf{Pool Size.}
The number of images in a pool $P$ to 2, 4, or 9.

\item \textbf{Image Domain.}
By default we use VQA images (\ie, from COCO~\cite{LinECCV14coco}),
but we also construct pools using CUB (bird) images~\cite{cub} and
AWA (animal) images~\cite{awa2}.

\item \textbf{Pool Sampling Strategy.}
We test two ways of sampling pools of images. The Constrast sampling method,
required for pre-training (\secref{sec:stage1}),
chooses a pair of images with contrasting answers to the same question from VQAv2.
This method only works for $P=2$.
The Random sampling method chooses $P$ images at random from the images available
in the split.
\end{compactitem}
For example, consider the `VQA - 2 Contrast - 5 Round' setting. 
These pools are constructed from 2 VQA images with the Contrast sampling strategy and dialogs are rolled out to 5 rounds.

\subsection{Baselines}
\label{sec:baselines}

We compare our proposed approach to two baselines -- \textbf{\stageone{}} and \textbf{\nonvar{}} -- ablating aspects of our model that promote
adaptation to new tasks or prevent language drift.
The \textbf{\stageone{}} baseline is our model after the single round fully supervised pre-training.
Improvements over this model represent gains made from task based fine-tuning.
The \textbf{\nonvar{}} baseline is our model under standard encoder-decoder dialog model design choices -- \ie,
a continuous latent variable, maximum likelihood pre-training, and fine-tuning the speaker module. 
Improvements over this model represent gains made from the modifications aimed at preventing language drift described in \secref{sec:new_qbot} --
specifically, the discrete latent variable, ELBO pre-training, and frozen speaker module.

\begin{figure}[!h]\footnotesize
    \centering
    \vspace{-10pt}
    \includegraphics[width=\textwidth]{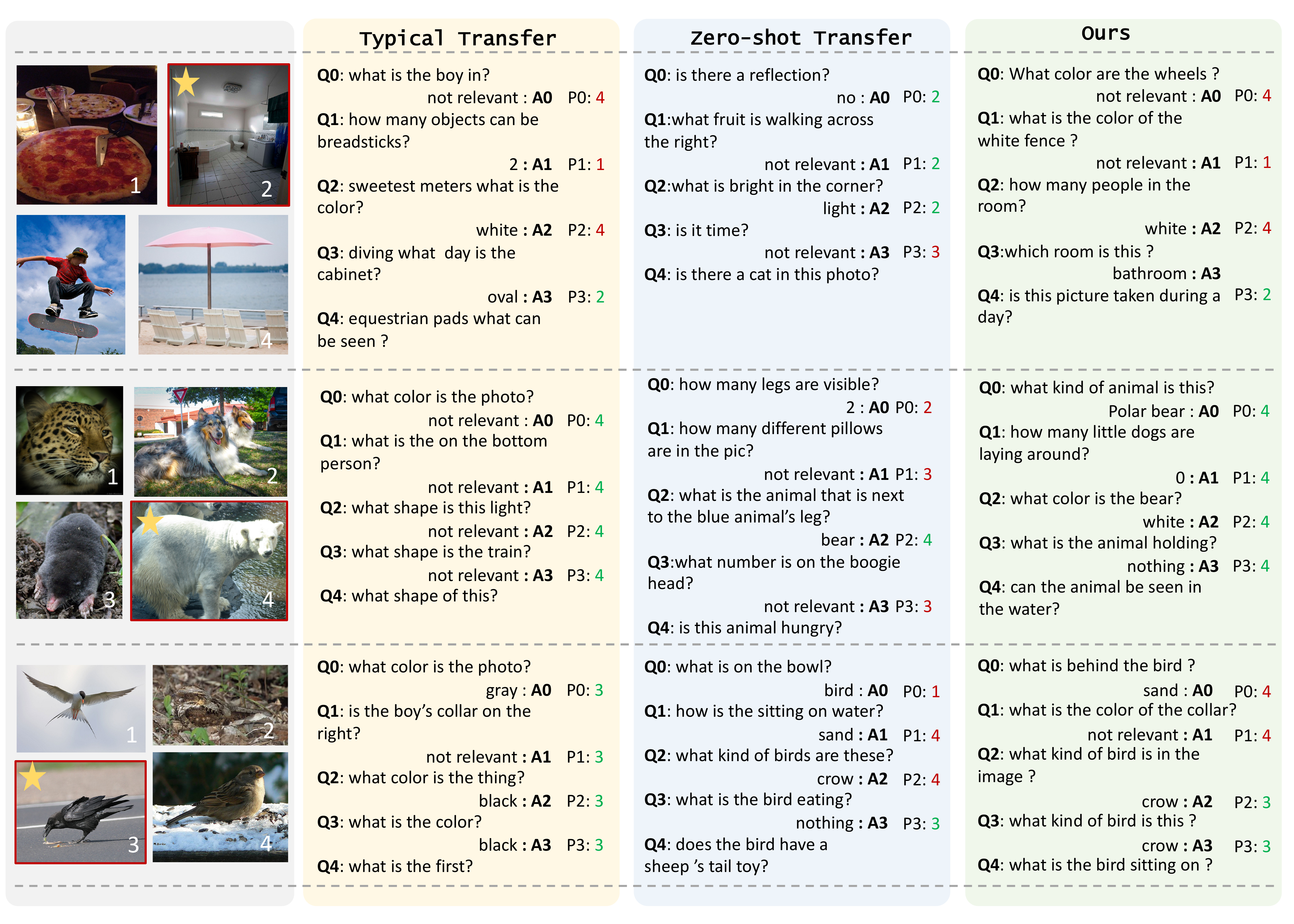}
    \vspace{-18pt}
    \caption{\footnotesize{Qualitative comparison of dialogs generated by our model with
        those generated by \nonvar{} and \stageone{} baselines. Top / middle / bottom rows are image pool from COCO / AWA / CUB images respectively. Our model is pre-trained on VQA (COCO images) and generates more intelligible questions on out-of-domain images.}}
    \vspace{-10pt}  
    \label{fig:examples}
\end{figure}

\subsection{Qualitative Results}
\label{sec:qualitative}
Figure~\ref{fig:examples} shows example outputs of the \nonvar{} and \stageone{} baselines alongside our \qbot{}
on VQA, AWA and CUB images using size 4 Randomly sampled pools and 5 rounds of dialog.
Both our model and the \nonvar{} baseline tend to guess the target image correctly,
but it is much easier to tell what the questions our model asks mean and how they might
help with guessing the target image.
On the other hand, questions from the \stageone{} baseline are clearly grounded in
the images, but they do not seem to help guess the target image and the \stageone{} baseline
indeed fails to guess correctly.
This is a pattern we will reinforce with quantitative results in \secref{sec:human-study} and \secref{sec:auto_eval}.

These examples and others we have observed suggest interesting patterns that highlight \abot{}.
Our automated \abot{} based on \cite{bottomup} does not always provide accurate answers,
limiting the questions \qbot{} can usefully ask.
When there is signal in the answers, it is not necessarily intelligible, providing an opportunity
for \qbot{}'s language to drift.

\subsection{Automated Evaluation}
\label{sec:auto_eval}


We consider metrics addressing both {\bf Task} performance and {\bf Language} quality.
While task performance is straightforward (did \qbot{} guess the correct target image?), language quality
is harder to measure. We describe three automated metrics here and further investigate language
quality using human evaluations in \secref{sec:human-study}.

\paragraph{Task -- Guessing Game Accuracy.}
To measure task performance so we report the accuracy of \qbot{}'s target image guess at the
final round of dialog.

\paragraph{Language -- Question Relevance via \abot{}.}
To be human understandable, the generated questions should be relevant to at least one image in the pool. We measure question relevance as the maximum question 
image relevance across the pool as measured by \abot{}, i.e., $1 - p(\texttt{Not Relevant})$. We note that this is only a proxy for actual question relevance as 
\abot{} may report \texttt{Not Relevant} erroneously if it fails to understand \qbot{}'s question; however, in practice we find \abot{} does a fair job in 
determining relevance. We also provide human relevance judgements in \secref{sec:human-study}. 



\paragraph{Language -- Fluency via Perplexity.}
To evaluate \qbot{}'s fluency, we train
an LSTM-based language model on the corpus of questions in VQA.
This allows us to evaluate the perplexity of the questions generated
by \qbot{} for dialogs on its new tasks. Lower perplexity indicates
the generated questions are similar to VQA questions in terms of syntax
and content. However, we note that questions
generated for the new tasks could have lower perplexity because they have drifted from English or because different
things must be asked for the new task, so lower perplexity
is not always better~\cite{gen_eval}.

\paragraph{Language -- Diversity via Distinct $n$-grams.}
This considers the set of all questions generated by \qbot{}
across all rounds of dialog on the val set.
It counts the number of $n$-grams in this set, $G_n$, and the number of
distinct $n$-grams in this set, $D_n$, then reports $\frac{G_n}{D_n}$
for each value of $n \in \{1, 2, 3, 4\}$.
Note that instead of normalizing by the number of words as in previous
work~\cite{dbs,li15ado}, we normalize by the number of n-grams so
that the metric represents a percentage for values of $n$ other than $n=1$.
Generative language models frequently produce safe standard outputs~\cite{dbs},
so diversity is a sign this problem is decreasing, but diversity by itself 
does not make language meaningful or useful.


\paragraph{Results.}
Table \ref{tab:main_results} presents results on our val set for our model and
baselines across the various settings described in \secref{sec:experimental_settings}.
Agents are tasked with generalizing further and further from their source language data.
Setting \texttt{A} is the same as for stage 1 pre-training.
In that same column, \texttt{B} and \texttt{C} 
require generalization to multiple rounds of dialog and Randomly sampled image pairs instead of pools sampled with the Contrast strategy.
In the right side of \tabref{tab:main_results} we continue to test generalization farther from the language source
using more images and rounds of dialog (\texttt{D}) and then using different types of images (\texttt{E} and \texttt{F}).
Our model performs well on both task performance and language quality across the different settings in terms of these
automatic evaluation metrics. Other notable findings are:
 
\begin{table}[h]
\newcommand{\g}{\cellcolor{gray!10}}
\newcommand{\num}[1]{\texttt{\footnotesize #1}}

\renewcommand*{\arraystretch}{1.25}
\resizebox{\columnwidth}{!}{
\begin{tabular}{c c l r r r r | c c l r r r r}
\toprule
 & & \small  & \small Accuracy $\uparrow$& \small Perplexity$\downarrow$ & \small Relevance $\uparrow$& \small Diversity$\uparrow$ & & & \small  & \small Accuracy $\uparrow$& \small Perplexity$\downarrow$ & \small Relevance $\uparrow$& \small Diversity$\uparrow$\\
 \midrule
 \multirow{3}{*}{\small \rotatebox{90}{\shortstack{\textbf{VQA} \\ 2 Contrast\\ 1 Round}}} &  
 \g \num{A1} & \g \stageone & \g0.73 &      \g2.62 &         \g0.87 &        \g0.50 & \multirow{3}{*}{\small \rotatebox{90}{\shortstack{\textbf{VQA} \\ 9 Random\\9 Rounds}}}
  & \g \num{D1} & \g \stageone & \g 0.18 &      \g 2.72 &         \g \textbf{0.77} &        \g 1.11 \\
  & \num{A2} & \nonvar & 0.71 &     10.62 &         0.66 &        \textbf{5.55} &  & \num{D2} & \nonvar &  \textbf{0.78} &     40.66 &       \textbf{0.77} &       \textbf{2.57} \\
  & \g\num{A3} & \g\ours & \g\textbf{0.82} & \g\textbf{2.6} & \g\textbf{0.88} & \g0.54 &   & \g \num{D3} & \g \ours & \g 0.53 &     \g \textbf{2.55} &       \g  0.75 &       \g 0.95 \\
\midrule
\multirow{3}{*}{\small \rotatebox{90}{\shortstack{\textbf{VQA} \\ 2 Contrast\\5 Rounds}}}
  & \num{B1} & \stageone & 0.67 &      2.62 &         0.87 &        0.50 & \multirow{3}{*}{\small \rotatebox{90}{\shortstack{\textbf{AWA} \\ 9 Random\\9 Rounds}}}
  & \num{E1} & \stageone &  0.47 &      2.49 &         \textbf{0.96} &        0.24 \\
  & \g\num{B2} & \g\nonvar & \g0.74 &     \g10.62 &         \g0.66 &       \g \textbf{5.55} &   & \g \num{E2} & \g \nonvar & \g 0.48 &     \g 12.56 &        \g 0.64 &        \g \textbf{2.21} \\

  & \num{B3} & \ours & \textbf{0.87} &      \textbf{2.60} &         \textbf{0.88} &        0.54 &   & \num{E3} & \ours & \textbf{0.74} & \textbf{2.41} &   \textbf{0.96} &  0.28 \\
\midrule
\multirow{3}{*}{\small \rotatebox{90}{\shortstack{\textbf{VQA} \\ 2 Random\\5 Rounds}}}
  & \g\num{C1} & \g\stageone & \g0.64 &    \g  \textbf{2.64} &     \g    0.75 & \g       1.73 & \multirow{3}{*}{\small \rotatebox{90}{\shortstack{\textbf{CUB} \\ 9 Random\\9 Rounds}}}
  & \g \num{F1} & \g \stageone & \g 0.36 &  \g    2.56 &         \g \textbf{1.00} &     \g   0.04 \\
  & \num{C2} & \nonvar & 0.86 &     16.95 &         0.62 &        \textbf{8.13} &   & \num{F2} & \nonvar &0.38 &     20.92 &         0.47 &        \textbf{2.16} \\
  &\g \num{C3} &\g \ours &\g \textbf{0.95} & \g     2.69 &    \g     \textbf{0.77} &     \g   2.34 &   &\g \num{F3} & \g \ours & \g \textbf{0.74} &     \g \textbf{2.47} &        \g \textbf{1.00} &    \g    0.04 \\
\bottomrule
\end{tabular}}

\caption{
Performance of our models and baselines in different experimental settings. From setting \texttt{A} to setting \texttt{F}, agents are tasked with generalizing further from the source data. Our method strikes a balance between guessing game performance and interpretability.
}
\vspace{-10pt}
\label{tab:main_results}
\end{table}

\paragraph{Ours \vs \stageone{}.} To understand the relative importance of the proposed stage 2 training which transferring to dialog for DwD task, we compared the task accuracy of our model with that of \stageone{}. In setting, \texttt{A} which matches the training regime, our model outperforms \stageone{} by 9\% (\texttt{A3} \vs \texttt{A1}) on task performance. As the tasks differ in settings B-F, we see further gains with our model consistently outperforming \stageone{} by 20-38\%. Despite these gains, our model maintains similar language perplexity, A-bot relevance, and diversity. 

\paragraph{Ours \vs \nonvar{}.} Our discrete latent variable, variational pre-training objective, and fixed speaker play an important role in avoiding language drift. Compared to the \nonvar{} model without these techniques, our model achieves over 4x (\texttt{A2} / \texttt{A3}) lower perplexity and 10-53\% better A-bot Relevance. Our model also improves in averaged accuracy, which means more interpretable language also improves the task performance. Note that \nonvar{} has 2-100x higher diversity compared to our model, which is consistent with the gibberish we observe from that model (\eg, in \figref{fig:examples}) and further suggests its language is drifting away from English.

\paragraph{Results from Game Variations.} We consider the following variations on the game:
\begin{compactitem}[--]
\item \textbf{Dialog Rounds.} Longer dialogs (more rounds) achieve better accuracy (A3 vs B3). 

\item \textbf{Pool Sampling Strategy.} As expected, Random pools are easier compared to Contrast pools (B3 vs C3 accuracy), however language fluency and
relevance drop on the Random pools (B3 vs C3 perplexity and a-bot relevance).  

\item \textbf{Image Source.} CUB and AWA pools are harder compared to COCO image domain (D3 vs E3 vs F3). Surprisingly, our models maintains similar perplexity and high a-bot relevance even on these out-of-domain image pools. The \stageone{} and \nonvar{} baselines generalize poorly to these different image domains -- reporting task accuracies nearly half our model performance.

\end{compactitem}

\subsection{Human Studies}
\label{sec:human-study}

We also evaluate our models by asking if humans can understand \qbot{}'s language.
Specifically, we use workers (turkers) on Amazon Mechanical Turk to evaluate the relevance, fluency,
and task performance of our models. Section \suppSectionHumanStudies{} of the supplement details these studies,
but we briefly summarize the results here.

Turkers from Amazon's Mechanical Turk considered the questions from our model more relevant to the image pool than those
from the \nonvar{} model and about equally as relevant as the \stageone{} model's questions.
Similarly, they considered our model's questions more fluent than the \nonvar{} model questions and equally
as fluent as the \stageone{} model's questions.
However, when we used used turkers to answer \qbot{}'s questions
-- replacing the automated \abot{} -- our \qbot{} was able to guess the correct image 69\% of the time while
the \nonvar{} only achieved 45\% accuracy and the \stageone{} model achieved 23\% accuracy.
This again confirms that our model can adapt to new tasks with minimal
sacrifice to language quality.

\subsection{Model Ablations}
\label{sec:ablations}

We investigate the impact of our modelling choices from \secref{sec:new_qbot}
by ablating these choices in Section \suppSectionAblations{} of the supplement, summarizing the
results here.
The choice of discrete instead of continuous $z^r$ helps maintain language
quality, as does the use of variational (ELBO) pre-training instead of maximum likelihood.
Surprisingly, the ELBO loss probably has more impact than the discreteness of $z^r$.
Fixing the speaker module during stage 2 also had a minor role in discouraging language drift.
Finally, we find that improvements in task performance are due more to learning to track
the dialog in stage 2.A than they are due to asking more discriminative questions.
\section{Related Work}
\label{sec:rel_work}

Our interest comes from language drift problems encountered when using models comparable to
the \stageone{} baseline. In \cite{deal_or_no_deal} a dataset
is collected with question supervision then fine-tuning is used in
an attempt to increase task performance, but the resulting utterances
are unintelligible. Similarly, \cite{visdial_rl} takes a very careful
approach to fine-tuning for task performance but finds that
language also diverges, becoming difficult for humans to understand.
Neither approach uses a discrete latent variables or a multi stage training curriculum, as in our proposed model.
Furthermore, these models need to be adapted to work in our new setting, and doing so would yield
models very similar to our \nonvar{} baseline.

More recently, \cite{drift_emnlp19} observe language drift in a translation game from French to German.
They reduce drift by supervising communications between agents with auxiliary translations to English and grounding in images.
This setting is somewhat different than ours since grounding is directly necessary to solve our task.
The approach also requires direct supervision on the communication channel, which is not practical
for a multiple round dialog game like ours.

We used a visual reference game to study question generation,
improving the quality of generated language using concepts related to latent action spaces.
Some works like \cite{lba} and \cite{visual_curiosity} also aim
to ask visual questions with limited question supervision.
Other works represent dialog using latent action spaces~\cite{rethink_zhao, zhao_unsup, zero_shot_zhao, yarats2017hierarchical, wen2017latent, serban2016piecewise, yarats2017hierarchical, hu2019hierarchical, kang2019recommendation, serban2017hierarchical, williams2017hybrid}. 
Finally, reference games are generally popular for studying language~\cite{lewis,guesswhat,visdial_eval}.
Section \suppSectionRelWork{} of the supplement describes the relationship between our approach
and these works in more detail.
\section{Conclusion}
\label{sec:conclusion}

In this paper we proposed the Dialog without Dialog (\dwd{}) task along with a model
designed to solve this task and an evaluation scheme that takes its goals into account.
The task is to perform the image guessing game, maintaining language quality without dialog level supervision.
This balance is hard to strike, but our proposed model manages to strike it.
Our model approaches this task by representing dialogs with a discrete latent variable and
carefully transfering language information via multi stage training.
While baseline models either adapt well to new tasks or maintain language
quality and intelligibility, our model is the only one to do both
according to both automated metrics and human judgements.
We hope these contributions help inspire useful dialog agents that can also interact with humans.

\section{Broader Impact}
\label{sec:broader_impact}

We think the main ethical aspects of this work and their consequences for society have to do with fairness. There is an open research problem around existing deep learning models often reflecting and amplifying undesirable biases that exist in society.

While visual question answering and visual dialog models do not currently work well enough to be relied on in the real world (largely because of the aforementioned proneness to bias), they could be deployed in applications where these biases could have negative impacts on fairness in the future. 
For example, visually impaired users might use these models to understand visual aspects of their world~\cite{vizwiz}. If these models are not familiar with people in certain contexts (e.g., men shopping) or are only used to interacting with certain users (e.g., native English speakers) then they might fail for some sub-groups (e.g., non-native English speakers who go shopping with men) but not others.

Our research model may be prone to biases, though it was trained on the VQAv2 dataset~\cite{vqav2}, which aimed to be more balanced than its predecessor. However, by increasing the intelligibility of generated language our work may help increase the overall interpretability of models. This may help by making bias easier to measure and providing additional avenues for correcting it.

i\section{Acknowledgements}

The Georgia Tech effort was supported in part by NSF, AFRL, DARPA, ONR YIPs, ARO PECASE, Amazon. The views and conclusions contained herein are those of the authors and should not be interpreted as necessarily representing the official policies or endorsements, either expressed or implied, of the U.S. Government, or any sponsor.

\bibliography{bib/main.bib}

\section{Supplement Overview}

This document contains supplementary material for ``Dialog without Dialog Data: Learning Visual Dialog Agents from VQA Data''.
The main paper excludes some details which we provide here.
\secref{sec:arch_details} describes the \qbot{} proposed in the main paper in more detail, including
algorithms that show how it executes one round of dialog.
\secref{sec:human_studies} describes the human studies we use to evaluate our model
and reports those results in detail.
\secref{sec:ablations} reports the ablations we use to evaluate the effects
of different aspects of the proposed \qbot{}.
\secref{sec:round_perf} reports how various models we consider perform at
different rounds of dialog.
Finally, \secref{sec:rel_work} explores in more depth how our work relates
to other relevant work in the literature.
\section{Architecture Details}
\label{sec:arch_details}

\SetKwProg{Fn}{Function}{}{end}
\SetKwFunction{QBot}{QBot}
\SetKwFunction{Planner}{Planner}
\SetKwFunction{Predictor}{Predictor}
\SetKwFunction{Speaker}{Speaker}
\SetKwFunction{Encoder}{Encoder}

This section describes our architecture in more detail.
\algref{alg:qbot} summarizes our complete \QBot{} implementation
and subsequent algorithms define the subroutines used inside \QBot{}
along with the encoder we use for variational pre-training.
The planner module is described in \algref{alg:planner},
the predictor is described in \algref{alg:predictor}, and
the speaker is described in \algref{alg:speaker}.
\algref{alg:encoder} describes the encoder used for the ELBO loss.

Note that the number of bounding boxes per image is $B$,
the number of images in a pool is $P$, and the
max question length is $T$.

There is a minor notation difference between this section and the main paper.
In this section there is an additional hidden state $\bar{h}_r$ that parallels
$h_r$ and is used only inside the planner. While $h_r$ is the hidden state of
an LSTM, $\bar{h}_r$ is computed in the same way except it uses a different
output gate (see line 11 of \algref{alg:planner}). This is essentially a second
LSTM output that allows the context coder query to forget dialog history
information irrelevant to the current round, and
allowing $h_r$ to focus on representing the entire dialog state.

\begin{algorithm}[h]
\Fn{\QBot{$\mathcal{I}, h_{r-1}, \bar{h}_{r-1}, q_0, a_0, \ldots, q_{r-1}, a_{r-1}$}}{
\KwIn{$\mathcal{I}, h_{r-1}, \bar{h}_{r-1}, q_0, a_0, \ldots, q_{r-1}, a_{r-1}$}
\KwOut{$q_{r}, h_{r}, \bar{h}_{r}, y_r$}
$y_r \gets$ \Predictor{$\mathcal{I}, h_{r-1}, q_0, a_0, \ldots, q_{r-1}, a_{r-1}$} \\
$h_{r}, \bar{h}_{r}, z^r \gets$ \Planner{$\mathcal{I}, q_{r-1}, a_{r-1}, h_{r-1}, \bar{h}_{r-1}$} \\
$q_{r} \gets$ \Speaker{$z^r$} \\
\Return $q_{r}, h_{r}, \bar{h}_{r}, y_r$ \\
}
\caption{Question Bot}
\label{alg:qbot}
\end{algorithm}

\begin{algorithm}[h]
\Fn{\Predictor{$\mathcal{I}, h_{r-1}, q_0, a_0, \ldots, q_{r-1}, a_{r-1}$}}{
\KwIn{$\mathcal{I}$ ($I_p^b \in \mathbb{R}^{2048}$), $h_{r-1}$, $q_0, a_0, \ldots, q_{r-1}, a_{r-1}$}
\KwOut{$y_r$}
$\mathrm{Attention}(Q, K, V) = \softmax{g_3(g_1(Q) \odot g_2(K))} V$ \\
$f_{r-1} \gets [E_q(q_{r-1}), E_a(a_{r-1})]$ \tcc{fact}
$F \gets [f_0, \ldots, f_{r-1}]$ \\
\tcc{Attention over rounds}
$e_F \gets \mathrm{Attention}(h_{r-1}, F, F)$ \\
$Q_y \gets [h_{r-1}, e_F]$ \\
\tcc{Attention over bounding boxes}
$e_I \gets \mathrm{Attention}(Q_y, {\bf x}, {\bf x}) \in \mathbb{R}^{P \times 2048}$ \\
$e_I \gets g_1(e_I)$ \\
$Q_p \gets g_2(Q_y)$ \\
$l_y \gets g_3(Q_p \odot e_I)$ \\
$y_r \gets \argmax{\softmax{l_y}}$ \\
\Return $y_r$ \\
}
\caption{Predictor}
\label{alg:predictor}
\end{algorithm}

\begin{algorithm}[h]
\Fn{\Planner{$\mathcal{I}, q_{r-1}, a_{r-1}, h_{r-1}, \bar{h}_{r-1}$}}{
\KwIn{$\mathcal{I}$ ($I^p_b \in \mathbb{R}^{2048}$), $q_{r-1}$, $a_{r-1}$, $h_{r-1}$, $\bar{h}_{r-1}$}
\KwOut{$h_{r}$, $\bar{h}_{r}$, $z^r$}
\tcc{Context Coder}
$e_q \gets E_q(q_{r-1})$ \\ 
$e_a \gets E_a(a_{r-1})$ \\ 
$e_c \gets f_5([\bar{h}_{r-1}, e_q, e_a])$ \\
$\alpha_p \gets \softmax{f_2(g(e_c) \odot f_1(I^p_b))}$ \\
$\beta^p_b \gets \softmax{f_4(g(e_c) \odot f_3(I^p_b))}$ \\
$\hat{v}_{r-1} \gets \sum_{p=1}^P \sum_{b=1}^B \alpha_p \beta^p_b I^p_b$ \\
$x^{context}_{r-1} \gets [\hat{v}_{r-1}, e_{q}, e_{a}]$ \\
\tcc{Dialog RNN}
$h_{r}, c_{r} \gets \gamma(x^{context}_{r-1}, h_{r-1})$ \\
$h_{r} \gets \mathrm{Dropout}(h_{r})$ \\
$\bar{h}_{r} \gets \sigma(W_1^T x^{context}_{r-1} + W_2^T h_{r-1}) \odot \tanh(c_{r})$ \\
$\bar{h}_{r} \gets \mathrm{Dropout}(\bar{h}_{r})$ \\
\tcc{Question Policy}
$l_{n,k} \gets \mathrm{LogSoftmax}\left(W^z_n h_{r} \right)_k \; \forall k \in \{1, \ldots, K\} \; \forall n \in \{1, \ldots, N\}$ \\
$z^r_n \gets \mathrm{GumbelSoftmax}(l_{n}) \; \forall n \in \{1, \ldots, N\}$ \\
$h_{r} \gets h_{r} + \relu{W^l l}$ \\
\Return $h_{r}, \bar{h}_{r}, z^r$
}
\caption{Planner}
\label{alg:planner}
\end{algorithm}

In the planner \algref{alg:planner}
at lines 5 and 6 $g, f_1, f_3$ are all two layer MLPs with ReLU output and weight norm.
Both $f_2$ and $f_4$ are linear transformations with weight norm applied (no activation function).
$f_5$ is a linear transformation without weight norm purely for dimensionality reduction.
To compute $\bar{h}_{r}$ we also add new linear weights $W_1$ and $W_2$ as for a standard LSTM output gate.

Note that for the planner there is an additional residual connection at line 16 which augments the
hidden state. This allows gradients to flow through the question policy parameters $W^z$ at line 13
when we fine-tune for task performance without fully supervised dialogs.

In \algref{alg:predictor} $g_1, g_2$ are both 2-layer ReLU nets with weight norm.
Also $g_3$ is a 2-layer net with ReLU and Dropout on the hidden activation and weight norm on both layers.

\begin{algorithm}[h]
\Fn{\Speaker{$z$}}{
\KwIn{$z$}
\KwOut{$q_{r+1}$}
$e_z \gets \sum_{n=0}^{N-1} E^z_n(z_n)$ \\
$q_{r+1} \gets \beta(e_z)$ \\
\Return $q_{r+1}$ \\
}
\caption{Speaker}
\label{alg:speaker}
\end{algorithm}

In \algref{alg:speaker} $\beta$ is an LSTM decoder.
\begin{algorithm}[h]
\Fn{\Encoder{$\mathcal{I}, q_r$}}{
\KwIn{$\mathcal{I} (I_p^b \in \mathbb{R}^{2048})$, $q_r$}
\KwOut{$z$ (sample or distribution parameters)}
\tcc{Context Coder}
$e_q \gets E_q(q_r)$ \\ 
$\alpha_p \gets \softmax{f_2(g(e_q) \odot f_1(I^{p}_b))}$ \\
$\beta^p_{b} \gets \softmax{f_4(g(e_q) \odot f_3(I^{p}_b))}$ \\
$\hat{v} \gets \sum_{p=1}^P \sum_{b=1}^B \alpha_p \beta^p_b I^{p}_b$ \\
$h \gets W_z^T \hat{v}$ \\
$l_{n,k} \gets \mathrm{LogSoftmax}\left(W^z_n h \right)_k \; \forall k \in \{1, \ldots, K\} \; \forall n \in \{1, \ldots, N\}$ \\
$z_n \gets \mathrm{GumbelSoftmax}(l_{n}) \; \forall n \in \{1, \ldots, N\}$ \\
\Return $z$
}
\caption{Encoder}
\label{alg:encoder}
\end{algorithm}

\section{Human Evaluation}
\label{sec:human_studies}

\begin{figure}[!h]
    \centering
    \includegraphics[width=0.95\textwidth]{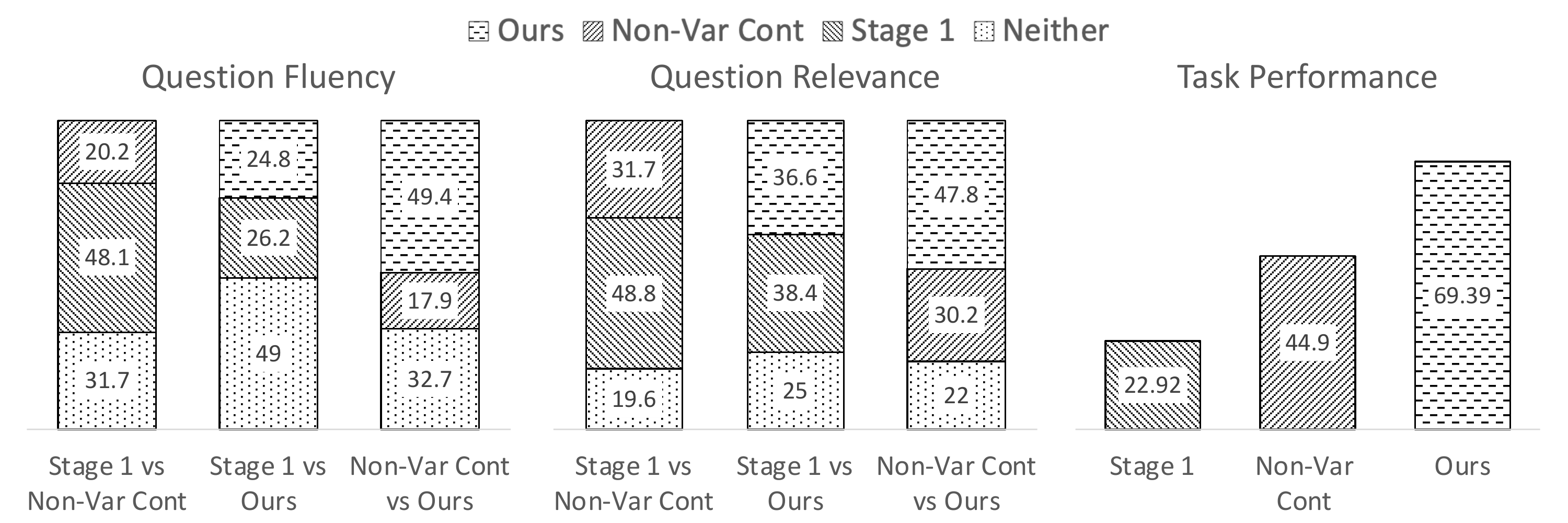}
\caption{Human evaluation of language quality -- question fluency (left), relevance (middle) and task performance (right). Question fluency and relevance compare a pair of agent-generated questions,
asking users which (or possibly neither) is more fluent/relevant. 
Task performance is to have humans interact dynamically with Q-bot in real time.
}
\label{fig:human_studies}
\end{figure}

As summarized in Section 4.5 of the main paper, we also evaluate our models by asking if humans can understand \qbot{}'s language.
Specifically, we use workers (turkers) on Amazon Mechanical Turk to evaluate the relevance, fluency,
and task performance of our models. We discuss each study below and report results for all studies in \figref{fig:human_studies}.

\begin{figure}[!h]
    \centering
    \includegraphics[width=\textwidth]{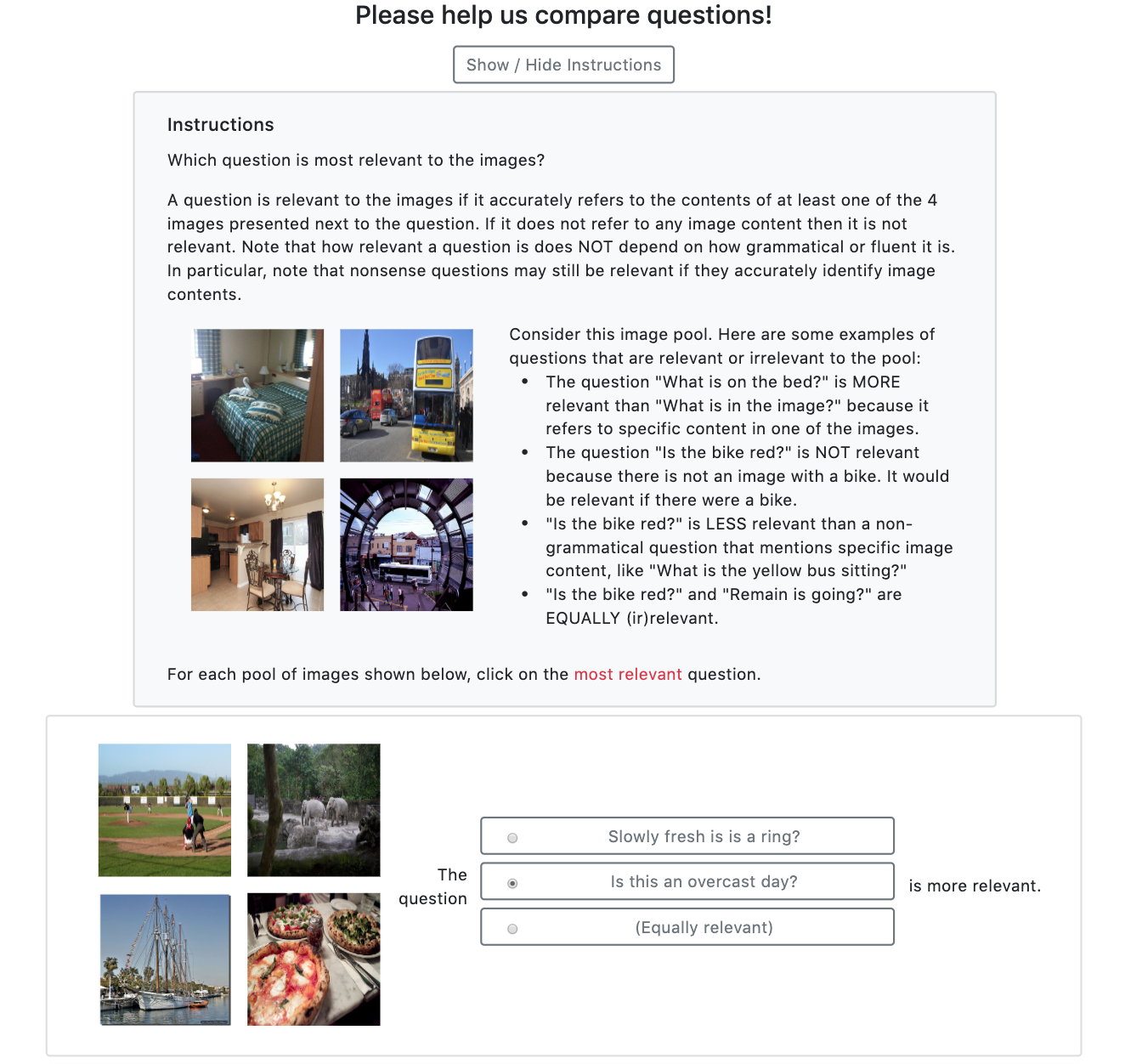}
    \caption{Human study instructions for question relevancy.}
    \label{fig:relevance_turk_interface}
\end{figure}

\paragraph{Human Study for Question Relevance.}
To get a more accurate measure of question relevance, we asked
humans to evaluate questions generated by our model and the baselines (\stageone{} \& \nonvar{}).
We curated 300 random, size 4 pools where all three models predicted the target correctly at round 5.
Size 9 pools require longer dialogs, so they take more effort for humans to analyze.
Humans can analyze more size 4 pools in the same time, so we use size 4 pools here.
For a random round, we show turkers the questions from a pair of models and ask \textit{`Which question is 
most relevant to the images?'}
Answering the question is a forced choice between three options:
one of the two models or an \textit{`Equally relevant'} option.
See \figref{fig:relevance_turk_interface} for an example of the interface we presented to turkers.
All model pairs were evaluated for each pool of images and
the questions were presented in a random order, though the \textit{`Equally relevant'} option was always last.

The results in \figref{fig:human_studies} (middle) show the frequency with which each
option was chosen for each model pair.
Our model was considered more relevant than the \nonvar{} model (47.8\% \vs 30.2\%)
and about the same as the \stageone{} model (36.6\% \vs 38.4\%). 

\begin{figure}[!h]
    \centering
    \includegraphics[width=\textwidth]{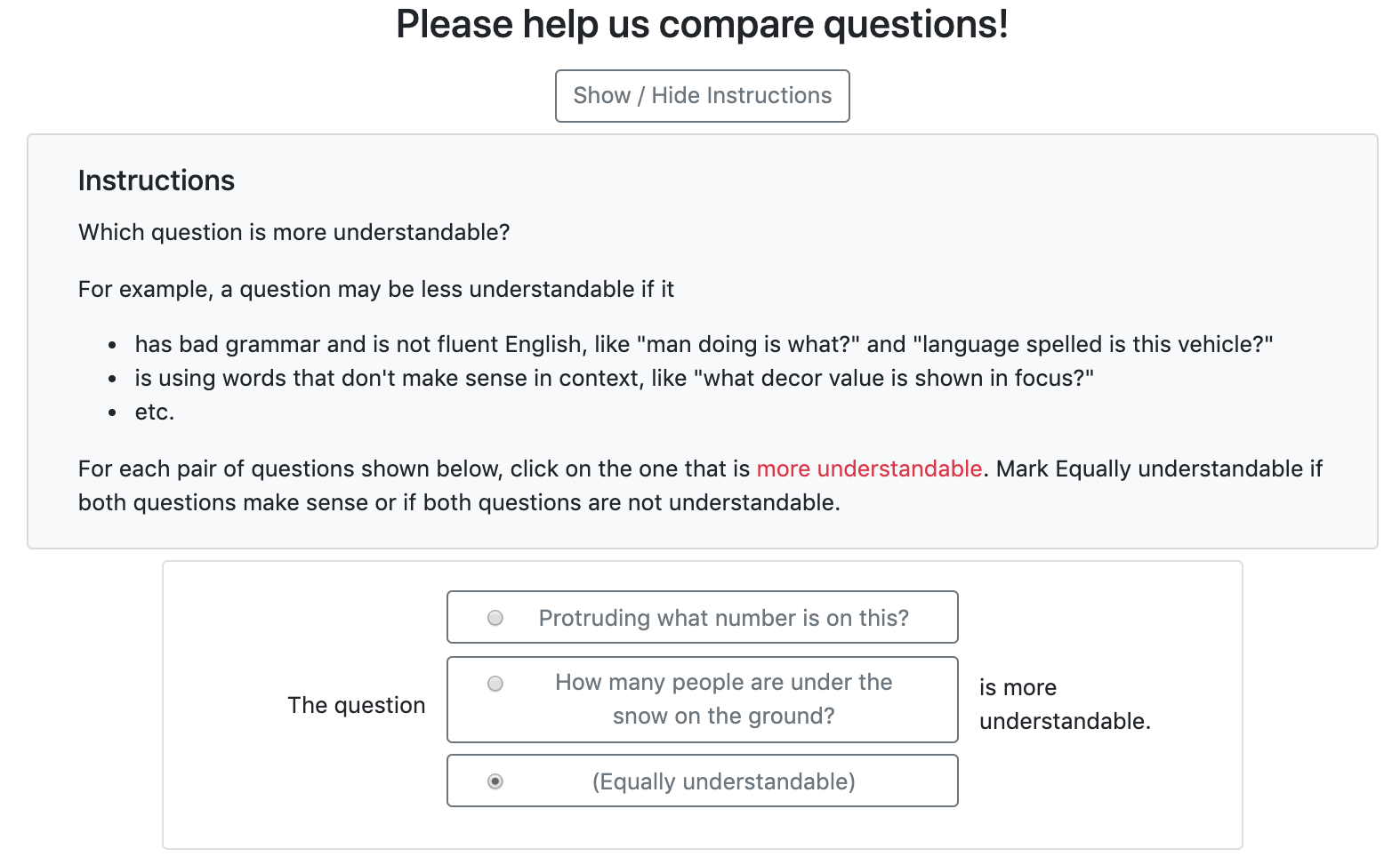}
    \caption{Human study instructions for question fluency.}
    \label{fig:fluency_turk_interface}
\end{figure}

\paragraph{Human Study for Question Fluency.}
We also evaluate fluency by asking humans to compare questions.
In particular, we presented the same pairs of questions to turkers
as in the relevance study, but this time we did \emph{not} present the pool
of images and asked them \textit{`Which question is more understandable?'}
As before, there was a forced choice between two models and an \textit{`Equally 
understandable'} option.
This captures fluency because humans are more likely to report that they understand
grammatical and fluent language.
An example interface is in \figref{fig:fluency_turk_interface}.
We used the same pairs of questions as in the relevance interface but
turkers were not given image pools with which to associate the questions.
As in the relevance study, questions were presented in a random order.

Figure \ref{fig:human_studies} (left) shows the frequency with which each
option was chosen for each model pair.
Our model is considered more fluent than the \nonvar{} model (49.4\% \vs 17.9\%) and about the same as 
the \stageone{} model (24.8\% \vs 26.2\%).

\begin{figure}[!h]
    \centering
    \includegraphics[width=\textwidth]{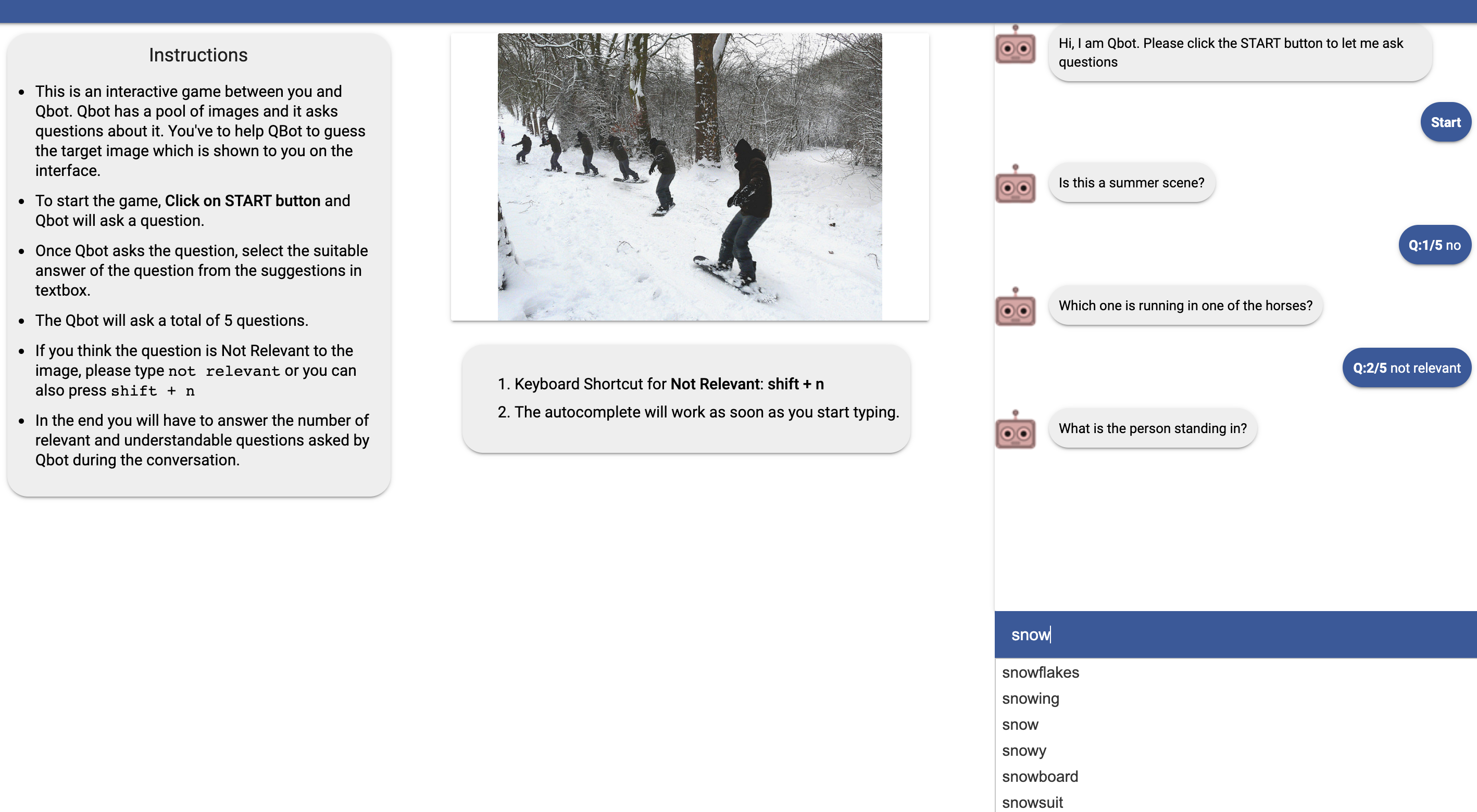}
    \caption{Interactive task performance MTurk interface. }
    \label{fig:interactive_turk_interface}
\end{figure}

\paragraph{Human Study for Task Performance.}
What we ultimately want in the long term is for humans to be able to collaborate with bots
to solve tasks. Therefore, the most direct evaluation of our the
\dwd{} task is to have humans interact dynamically with \qbot{}.
We implemented an interface that allowed turkers to interact with
\qbot{} in real time, depicted in \figref{fig:interactive_turk_interface}.
\qbot{} asks a question. A human answers it after looking at the target image.
\qbot{} asks a new question in response to the human answer and the
human responds to that question. After the 4th answer \qbot{}
makes a guess about which target image the human was answering based on.

We perform this study for the same pools for each model and find our approach achieves an accuracy of 69.39\% -- significantly higher
than \nonvar{} at 44.90\% and \stageone{} at 22.92\% as shown in \figref{fig:human_studies} (right). This study shows that our model
learns language for this task that is amenable to human-AI collaboration. This is in contrast to prior work \cite{visdial_eval} that
showed that improvements captured by task-trained models for similar image-retrieval tasks did not transfer when paired with human partners.

\section{Model Ablations}
\label{sec:ablations}

We investigate the impact of our modelling choices from Section 3.2 of the main paper.
In \tabref{tab:ablations} we report the mean
of all four automated metrics averaged
over pool sizes, pool sampling strategies, and datasets.\footnotemark
Next we explain how we vary each of these model dimensions
\footnotetext{This includes 10 settings: 
\{random 2, 4, 9 pools
\}$\times$ \{VQA, AWA, CUB\} and 2 contrats pools on VQA}





\begin{compactitem}[\hspace{3pt}--]
\item
Our 128 4-way Concrete variables $z^r_1, \ldots, z^r_{128}$ require 512 logits ({\bf Discrete}).
Thus we compare to the standard Gaussian random variable
common throughout VAEs with 512 dimensions ({\bf Continuous}).

\item
In both discrete and continuous cases we train with an ELBO
loss ({\bf ELBO}), so we compare to a maximum likelihood only
model ({\bf MLE}) that uses an identity function 
as in the default option for the Question Policy (see Section 2.3.1 of the main paper).
The MLE model essentially removes the KL term (2nd term of Eq. 7 of the main paper)
and ignores the encoder during pre-training.

\item
We consider checkpoints after each step of our training curriculum: {\bf Stage 1}, {\bf Stage 2.A}, and {\bf Stage 2.B}.
For some approaches we skip Stage 2.A and go straight to fine-tuning everything except the speaker as in Stage 2.B. This is denoted by {\bf Stage 2}.

\item
We consider 3 variations on how the speaker is fine-tuned.
The first is our proposed approach of fixing the speaker ({\bf Fixed}).
The next fine-tunes the speaker ({\bf Fine-tuned}).
To evaluate the impact of fine-tuning we also consider
a version of the speaker which can not learn to ask better questions
by using a parallel version of the same model ({\bf Parallel}).
This last version will be described more below.
\end{compactitem}

\begin{table}

{
\newcommand{\num}[1]{\texttt{\footnotesize #1}}
\newcommand{\g}{\cellcolor{gray!10}}

\resizebox{\columnwidth}{!}{
\begin{tabular}{l l l l l c c c c}
\toprule
 & $z$ Structure & Loss & Curriculum & Speaker & Accuracy & Perplexity & Relevance & Diversity \\
\midrule
\g \num{1} & \g Discrete &\g ELBO & \g Stage 2.B & \g Fixed  &     \g 0.81 &      \g 2.57 &        \g 0.89 &       \g \textbf{0.86} \\
\num{2} &Discrete &ELBO & Stage 2 & Fine-tuned          &     \textbf{0.82} &      2.54 &         0.85 &        0.59 \\
\g \num{3} &\g Discrete &\g ELBO & \g Stage 2 &\g Parallel           &   \g  0.78 &    \g  2.60 &      \g   0.88 &     \g   0.73 \\
\num{4} &Discrete &ELBO & Stage 1 &Fixed           &     0.72 &      2.60 &         \textbf{0.91} &        0.48 \\
\g \num{5} &\g Discrete &\g ELBO & \g Stage 2.A & \g Fixed         &  \g   0.80 &    \g  2.59 &      \g   0.89 &    \g    0.81 \\
\num{6} &Discrete &ELBO & Stage 2 & Fixed           &     0.80 &      \textbf{2.53} &         0.85 &        0.62 \\
\g \num{7} &\g Continuous &\g ELBO & \g Stage 2.B &\g  Fixed         &   \g  0.75 &   \g  2.45 &      \g   0.66 &     \g   0.23 \\
\num{8} &Continuous & MLE & Stage 2.B & Fixed         &     0.78 &      4.27 &         0.83 &        4.33 \\
\bottomrule
\end{tabular}
}
}

\caption{Various ablations of our training curriculum.}
\label{tab:ablations}
\end{table}

\paragraph{Discrete Outperforms Continuous $z^r$.}
By comparing our model in row 1 of \tabref{tab:ablations} to row
7 we see that our discrete model outperforms the corresponding
continuous model in terms of task performance (higher Accuracy) and about matches it in interpretability (similar Perplexity and higher Relevance).
This may be a result of discreteness constraining the optimization
problem to prevent overfitting and is consistent with previous
work that used a discrete latent variable to model dialog~\cite{rethink_zhao}.

\paragraph{Stage 2.B Less Important than Stage 2.A}
Comparing rows 4, 5, and 1 of \tabref{tab:ablations}, we can see that
each additional step, Stage 2.A (row 4 \texttt{->} 5) and Stage 2.B (row 5 \texttt{->} 1),
increases task performance and stays about the same in terms of 
interpretability. However, most gains in task performance happen between Stage 1
and Stage 2. This indicates that improvements in task performance
are mainly from learning to incorporate information over multiple
rounds of dialog.

\paragraph{Better Predictions, Slightly Better Questions}
To further investigate whether \qbot{} is asking better questions
or just understanding dialog context for prediction better
we considered a {\bf Parallel} speaker model. This model
loaded two copies of \qbot{}, A and B both starting at the model resulting from Stage 1.
Copy A was fine-tuned for task performance, but every $z^r$ it generated
was ignored and replaced with the $z^r$ generated by copy B,
which was not updated at all.
The result was that copy A of the model could not incorporate dialog
context into its questions any better than the Stage 1 model,
so all it could do was track the dialog better for prediction
purposes.
By comparing the performance of copy A (row 3 of \tabref{tab:ablations})
to our model (row 1) we can see a 3 point different in accuracy, so
the question content of our model has improved after fine-tuning, but
not by a lot. Again, this indicates most improvements are from dialog tracking for prediction (row 3 accuracy
is much higher than row 4 accuracy).

\paragraph{Fine-tuned Speaker}
During both Stage 2.A and Stage 2.B we fix the Speaker module
because it is intended to capture low level language details and we
do not want it to change its understanding of English.
Row 2 of \tabref{tab:ablations} does not fix the Speaker during Stage 2 fine-tuning.
Instead, it uses each softmax at each step of the LSTM decoder to parameterize 
one Concrete variable~\cite{gumbel_softmax} per word. This allows gradients to
flow through the decoder during fine-tuning, allowing the model
to tune low-level signals. This is similar to previous approaches which
either used this technique~\cite{best_of_both_worlds} or REINFORCE~\cite{visdial_rl}
This model is competitive with \dwd{} in terms of task performance. However, when we inspect
its output we see somewhat less understandable language.

\paragraph{Variational Prior Helps Interpretability}
We found the most important factor for maintaining interpretability
to be the ELBO loss we applied during pre-training.
Comparing the continuous Gaussian variable (row 7) to a similar hidden
state (row 8) trained without the KL prior term we see 
drastically different perplexity and diversity. In the main paper these
metrics dropped when a model had drifted from English (\eg, for \nonvar{}).
This suggests the model without the ELBO in row 8 has experienced similar language drift.

\section{Performance by Round}
\label{sec:round_perf}

%

\begin{figure}[!h]
    \centering
    \includegraphics[width=\textwidth]{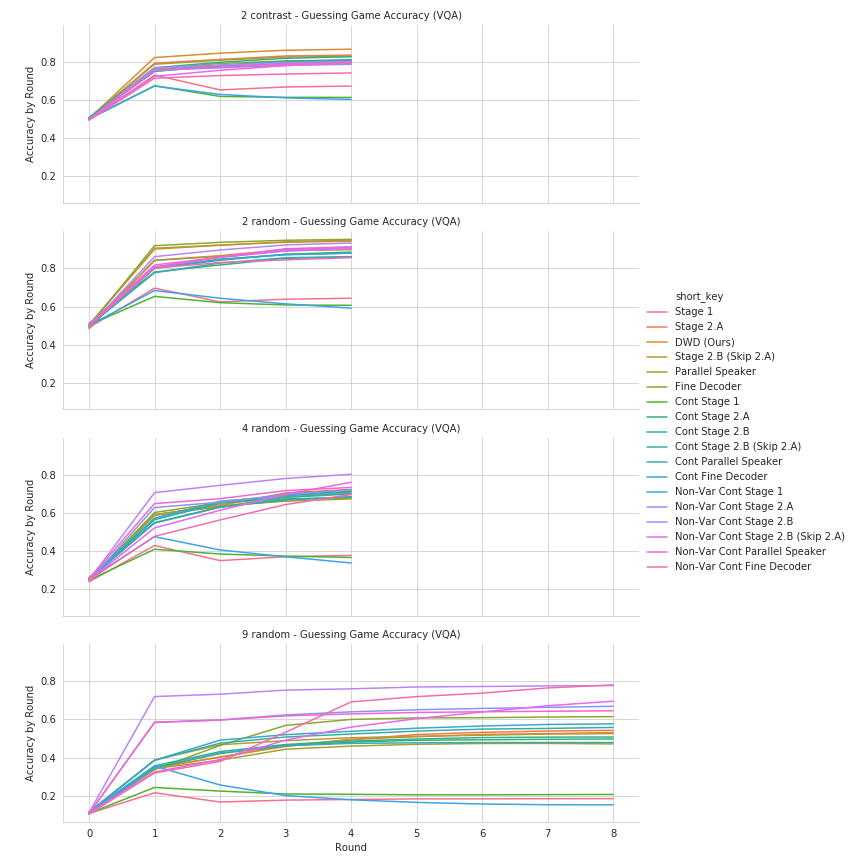}
    \caption{Task performance (guessing game accuracy) over rounds of dialog.
    Performance increases over rounds for all models except the Stage 1 models.}
    \label{fig:acc_over_rounds}
\end{figure}

Experiments in the main paper considered dialog performance after the first round (top of Table 1)
and at the final round of dialog (either 5 or 9).
This does not give much sense for how dialog performance increases over rounds of dialog,
so we report \QBot{}'s guessing game performance at each round of dialog in \figref{fig:acc_over_rounds}.
For all fine-tuned models performance goes up over multiple rounds of dialog, though some models
benefit more than others.
Stage 1 models decrease in performance after round 1 because it is too far from the training
data such models have been exposed to.

\section{Related Work}
\label{sec:rel_work}

We used a visual reference game to study question generation,
improving the quality of generated language using concepts related to latent action spaces.
This interest is mainly inspired by
problems encountered when using models comparable to the \stageone{} baseline. In \cite{deal_or_no_deal} a dataset
is collected with question supervision then fine-tuning is used in
an attempt to increase task performance, but the resulting utterances
are not intelligible. Similarly, \cite{visdial_rl} takes a very careful
approach to fine-tuning for task performance but finds that
language also diverges, becoming difficult for humans to understand.

\paragraph{Visual Question Generation.}
Other approaches like \cite{lba} and \cite{visual_curiosity} also aim
to ask questions with limited question supervision.
They give Q-bot access to an oracle to which it can ask
any question and get a good answer back.
This feedback allows these models to ask questions that
are more useful for teaching \abot{}~\cite{lba} or generating scene
graphs~\cite{visual_curiosity}, but they require a domain specific oracle
and do not take any measures to encourage interpretability.
We are also interested in generalizing with limited supervision,
using a standard VQAv2~\cite{vqav2} trained \abot{} as a flawed oracle,
but we focus on maintaining interpretability of generated questions and
not just their usefulness.

\paragraph{Latent Action Spaces.}
Of particular interest to us is a line of work that uses represents dialogs
using latent action spaces \cite{zhao_unsup, zero_shot_zhao, yarats2017hierarchical, wen2017latent, serban2016piecewise, yarats2017hierarchical, hu2019hierarchical, kang2019recommendation, serban2017hierarchical, williams2017hybrid}. 
Recent work uses these representations to discover intelligible language~\cite{zhao_unsup} and
to perform zero-shot dialog generation~\cite{zero_shot_zhao}, though neither works consider visually grounded language as in our approach. 
Most relevant is \cite{rethink_zhao}, which focuses on the difference between
word level feedback and latent action level feedback. Like us, they use
a variationally constrained latent action space (like our $z$) to generate
dialogs and find that by providing feedback to the latent actions instead of
the generated words (as opposed to the approaches in \cite{visdial_rl} and \cite{deal_or_no_deal}) they achieve better dialog performance.
Our variational prior is similar to the Full ELBO considered in
\cite{rethink_zhao}, but we consider generalization from
non-dialog data and generalization to new modalities.

\paragraph{Reference Games.}
The task we use to study question generation follows a body
of work that uses reference games to study language and its interaction
with other modalities~\cite{lewis}.
Our particular task is most similar to those in \cite{guesswhat} and \cite{visdial_eval}.
In particular, \cite{guesswhat} collects a dataset for goal oriented
visual dialog using a similar image reference game
and \cite{visdial_eval} uses a similar guessing game we use
to evaluate how well humans can interact with A-bot.

\end{document}